\documentclass[letterpaper, 10 pt, conference]{ieeeconf}  

\IEEEoverridecommandlockouts                              

\usepackage{siunitx}
\usepackage{graphics} 
\usepackage{graphicx}
\usepackage{placeins}

\usepackage{epsfig} 
\usepackage{amsmath} 
\usepackage{amssymb}  

\usepackage{bm}
\usepackage{mathtools}

\usepackage{algorithm}
\usepackage[noend]{algpseudocode} 
\algnewcommand\AAND{\textbf{ and }}
\algnewcommand\Or{\textbf{ or }}

\usepackage{color}
\usepackage{citesort}
\usepackage{flushend}
\usepackage{url}
\usepackage[breaklinks]{hyperref}
\usepackage[capitalise]{cleveref}
\usepackage{booktabs}
\usepackage{makecell}
\usepackage{multirow}
\usepackage[nolist,nohyperlinks]{acronym}
\acrodef{method}[AOM]{ACRONYM OF METHOD}
\acrodef{gnss}[GNSS]{Global Navigation Satellite System}
\acrodef{ransac}[RANSAC]{Random Sample Consensus}
\acrodef{slam}[SLAM]{Simultaneous Localization And Mapping}
\acrodef{pca}[PCA]{Principal Component Analysis}
\acrodef{ekf}[EKF]{Extended Kalman Filter}
\acrodef{rmse}[RMSE]{Root Mean Square Error} 
\acrodef{ape}[APE]{Absolute Pose Error}
\acrodef{cfar}[CFAR]{Constant False Alarm Rate}
\acrodef{snr}[SNR]{Signal to Noise Ratio}
\acrodef{rcs}[RCS]{Radar Cross Section}
\acrodef{imu}[IMU]{Inertial Measurement Unit}
\acrodef{sgm}[SGM]{Segmi-Global Matching}
\acrodef{dnn}[DNN]{Deep Neural Network}
\acrodef{gru}[GRU]{Gated Recurrent Unit}
\acrodef{hpr}[HPR]{Hidden Point Removal}
\acrodef{raft}[RAFT]{Recurrent All-Pairs Field Transforms}
\acrodef{fov}[FOV]{Field of View}
\acrodef{mclab}[MC-lab]{Marine Cybernetics laboratory}
\acrodef{vio}[VIO]{Visual-Inertial Odometry}
\acrodef{rcm}[RCM]{Refractive Camera Model}
\acrodef{sfm}[SFM]{Structure from Motion}
\acrodef{cnn}[CNN]{Convolutional Neural Network}
\acrodef{mse}[MSE]{Mean Squared Error}
\acrodef{gnll}[NLL]{Negative Log Likelihood}
\acrodef{rov}[ROV]{Remotely Operated Vehicle}
\acrodef{rovs}[ROVs]{Remotely Operated Vehicles}
\acrodef{deepvl}[DeepVL]{Deep Velocity Learning}

\acrodef{rpe}[RPE]{Relative Position Error}
\acrodef{dvl}[DVL]{Doppler Velocity Log}
\acrodef{reaqrovio}[ReAqROVIO]{Refractive Aquatic ROVIO}
\acrodef{deepvl}[DeepVL]{Deep Velocity Learning}
\acrodef{ve}[VE]{Volumetric Exploration}
\acrodef{gvi}[GVI]{General Visual Inspection}
\acrodef{tsp}[TSP]{Traveling Salesman Problem}
\acrodef{dof}[DoF]{Degree of Freedom}

\usepackage{amsmath}
\usepackage{tabularray}

\usepackage{tikz}
\usetikzlibrary{positioning, shapes.geometric, arrows.meta, decorations.pathreplacing, calligraphy}

\DeclareMathAlphabet{\pazocal}{OMS}{zplm}{m}{n}

\newcommand{\Ys}{\pazocal{Y}}

\DeclareMathAlphabet{\mathpzc}{OT1}{pzc}{m}{it}

\usepackage{array}
\newcolumntype{C}[1]{>{\centering\arraybackslash}p{#1}}
\newcolumntype{M}[1]{>{\raggedright\arraybackslash}p{#1}}

\usepackage{array} 
\newcolumntype{L}[1]{>{\raggedright\let\newline\\\arraybackslash\hspace{0pt}}m{#1}}	
\newcolumntype{S}[1]{>{\centering\let\newline\\\arraybackslash\hspace{0pt}}m{#1}}
\newcolumntype{R}[1]{>{\raggedleft\let\newline\\\arraybackslash\hspace{0pt}}m{#1}}

\makeatletter
\renewcommand*{\@opargbegintheorem}[3]{\trivlist
  \item[\hskip \labelsep{\itshape #1\ #2}] \textit{(#3)}\ }
\makeatother

\usepackage{fancyhdr}
\usepackage{eso-pic}
\fancypagestyle{withfooter}{
  
  \fancyhead[L]{}
  \fancyhead[R]{}
  \fancyfoot[C]{\footnotesize Presented at the 2025 IEEE ICRA Workshop on Field Robotics}
}

\DeclareMathAlphabet{\pazocal}{OMS}{zplm}{m}{n}

\title{\LARGE \bf
Ariel Explores: Vision-based underwater exploration and inspection via generalist drone-level autonomy
}

\author{Mohit Singh, Mihir Dharmadhikari and Kostas Alexis 
\thanks{This material was supported by the Research Council of Norway Awards NO-327292 and NO-321235.}
\thanks{The authors are with the Norwegian University of Science and Technology (NTNU), O. S. Bragstads Plass 2D, 7034, Trondheim, Norway {\tt\small mohit.singh@ntnu.no}}
}

\begin{document}

\maketitle
\thispagestyle{withfooter}
\pagestyle{withfooter}


\begin{abstract}
This work presents a vision-based underwater exploration and inspection autonomy solution integrated into Ariel, a custom vision-driven underwater robot. Ariel carries a $5$ camera and IMU based sensing suite, enabling a refraction-aware multi-camera visual-inertial state estimation method aided by a learning-based proprioceptive robot velocity prediction method that enhances robustness against visual degradation. Furthermore, our previously developed and extensively field-verified autonomous exploration and general visual inspection solution is integrated on Ariel, providing aerial drone-level autonomy underwater. The proposed system is field-tested in a submarine dry dock in Trondheim under challenging visual conditions. The field demonstration shows the robustness of the state estimation solution and the generalizability of the path planning techniques across robot embodiments. The video for our experiments is released at: \href{https://ntnu-arl.github.io/ariel-explores/}{https://ntnu-arl.github.io/ariel-explores/}
\end{abstract}

\section{INTRODUCTION}
The inspection of underwater structures is critical in a variety of domains, ranging from industrial infrastructure, such as offshore energy generation plants, ports, and aquaculture, to natural environments and sea life monitoring. Many of these environments are hard to access for humans and, at times, extremely hazardous. To tackle this challenge, underwater robotic solutions are employed to perform the inspection tasks. However, these operations are predominantly restricted to manually teleoperated robotic inspections requiring a tethered robot and an expert human pilot at the inspection site. 

The robotics community has contributed towards autonomous underwater robotic operations. These systems typically rely on domain specific sensing such as 3D sonars, acoustics, and Doppler Velocity Log~\cite{wu2019survey}, whereas vision cameras are primarily used for observation and less for autonomy. Some works have contributed to the field of autonomous exploration and inspection using acoustic sensors~\cite{2018vidaluwexp,Sureshactiveslam2020volexplration,JACOBI2015uwinsp}. In recent years, as the requirement for closeup inspection of structures and operation in confined environments has increased, vision-based underwater systems have gained popularity \cite{abdullah2024caveseg,hong2019water}. Several researchers have presented work in vision-driven underwater odometry methods \cite{rahman2022svin2,miao2021univio,joshi2023sm}. Despite the progress, such works largely remain tested in simple environments. Additionally, vision-driven underwater systems face challenges such as varying physical conditions, low visibility, low-illumination, etc.

\begin{figure}
    \centering
    \includegraphics[width=1\linewidth]{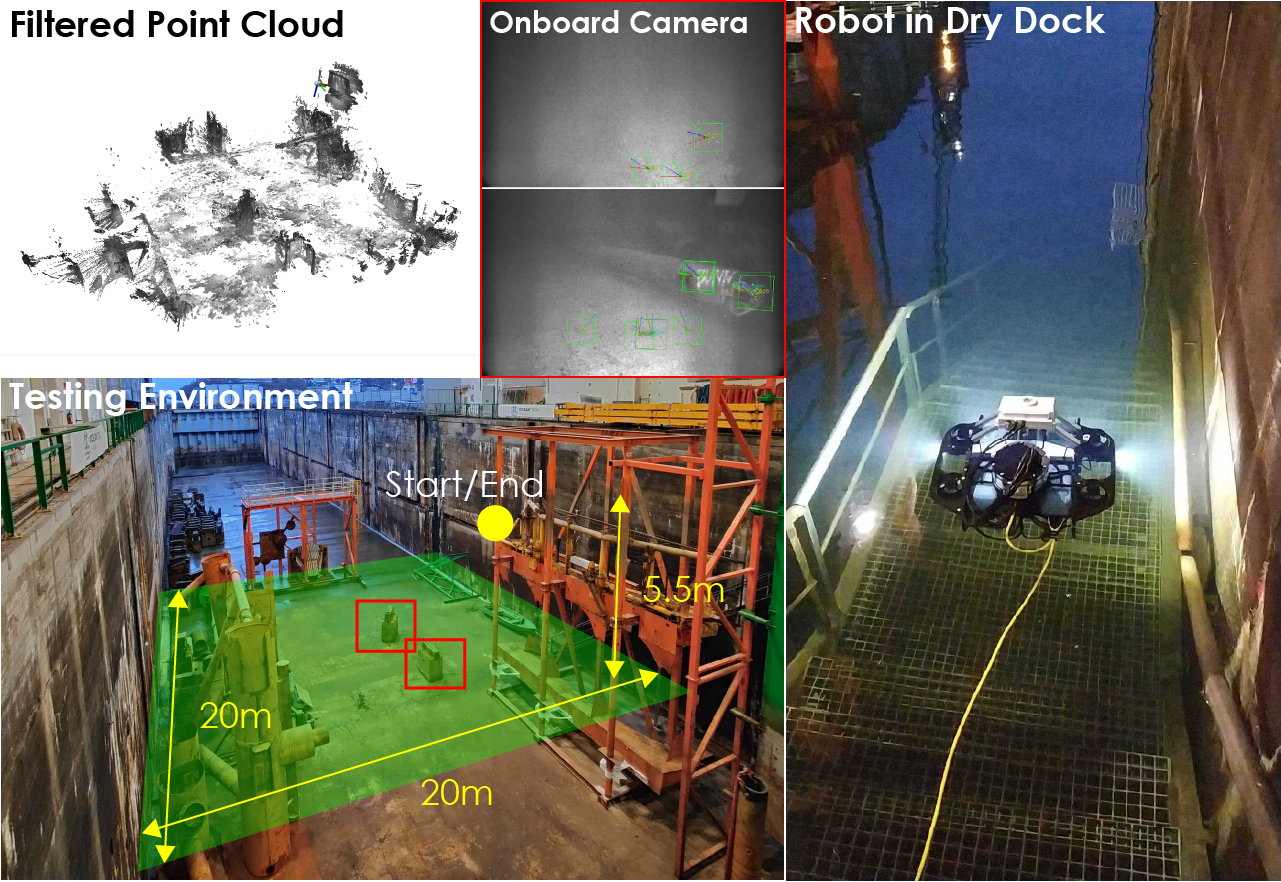}
    \caption{Instance of the robot in the dry dock. The subfigure on bottom left shows view of the empty dry dock showcasing the dimensions and the submerged structures. On the top left, the filtered point cloud map from the mission is shown along with instances of onboard camera images showing the low visibility conditions.}
    \label{fig:intro}
\end{figure}

Motivated by the above, in this paper, we present vision-based autonomous underwater exploration in visually degraded environments. We utilized our previous work on underwater refraction-aware visual inertial odometry \ac{reaqrovio}~\cite{SinghRCMinRovio2024}, which enables robust visual odometry while only needing calibration in air and estimating the refractive index to enable odometry underwater. It is further aided by \ac{deepvl}~\cite{singh2025deepvldynamicsinertialmeasurementsbased} a deep learning framework to predict robot velocity based on learned dynamics and inertial measurements, allowing long-term reliable state estimation during the lack or critical absence of reliable visual features in the environment. Furthermore, the robot exploits the methods described in~\cite{2023expgvi} for efficient exploration and inspection planning. The proposed robotic system is deployed in a submarine bunker dry dock under low visibility conditions~\ref {fig:intro}.

The remainder of the paper is organized as follows. Section~\ref{sec:vision} describes the visual inertial state estimation system, followed by vision-based mapping in Section~\ref{sec:mapping}. The exploration and inspection planning pipeline is detailed in Section~\ref{sec:exploration}, and the robotic system and field deployment is described in Section~\ref{sec:evaluation}. Finally, the lessons learned are discussed in Section~\ref{sec:lessons}, and conclusions are drawn in Section~\ref{sec:concl}.

\vfill


\section{Visual Inertial State Estimation}\label{sec:vision}
\subsection{Refractive Aquatic ROVIO}
We use \ac{reaqrovio}\cite{SinghRCMinRovio2024}, a refraction-aware multi-camera Visual-Inertial Odometry solution based on ROVIO \cite{bloesch2015robust}. It uses an Iterated Extended Kalman Filter as the back-end with \ac{imu} based motion model as the prediction model and tracks multilevel patch features (across multiple pyramid levels of the image) to enable state estimation.  In order to track the patches, it uses a projection function that models the projection of a given point in the environment on to the image coordinates. The projection function also uses refractive index as an argument.  \ac{reaqrovio} estimates the refractive index of the given media online as a part of the state of the underlying Iterated Extended Kalman Filter. It only requires camera calibration (extrisics and intrinsics) in air, and it enables reliable state estimation in a refractive media (water in present case) without needing to calibrate the cameras in a particular medium. This results in direct deployment of the robot in waters with drastically different physical and optical conditions (temperature, salinity, and depth). Furthermore, the multilevel patch feature formulation in \ac{reaqrovio} (adopted from ROVIO) enables reliable state estimation in the scenarios of weak image gradients alongside \ac{imu} prediction model-based robust feature tracking. \ac{reaqrovio} also fuses barometric pressure as a depth measurement update and includes an optional fusion of a velocity measurement update with covariance, we utilize the velocity fusion to further enhance the robustness by velocity aiding as described in next subsection.
 
\subsection{Deep Velocity Learning}
Deep Velocity learning (\ac{deepvl}) \cite{singh2025deepvldynamicsinertialmeasurementsbased} is a learned robot velocity prediction method that aids \ac{reaqrovio}, enabling state-estimation in low or complete absence of visual features. Primarily, it uses a network trained to output the velocity and associated covariance which are used as a velocity update in \ac{reaqrovio}. It contains a recurrent neural network based on an ensemble of \ac{gru} and takes the motor commands, \ac{imu} measurements, and battery voltage as input in a temporally recurrent manner. The network is trained using supervised learning to predict the robocentric body frame linear velocity and the corresponding uncertainty. The use of an ensemble of networks further enhances the uncertainty prediction. The network is trained on a dataset collected by manually piloting the robot for \SI{4}{h} in a pool and about \SI{20}{min} in the Trondheim Fjord. The predicted robocentric velocity and the uncertainty (as a covariance matrix) are used as a measurement update in \ac{reaqrovio}. The method provides reliable odometry in long-term visual blackout with \ac{rpe} \ac{rmse} error of \SI{0.2}{m} (delta of \SI{10}{m}) in controlled lab environments and \SI{0.4}{m} (delta of \SI{10}{m}) in experiments performed in the Trondheim fjord with trajectory lengths varying from \SI{122}{m} to \SI{305}{m}. Further details on the method and its evaluation can be found in~\cite{singh2025deepvldynamicsinertialmeasurementsbased}.

\section{Vision Based Mapping}\label{sec:mapping}
\subsection{Learning-based Stereo Vision Depth}
We use Raft-Stereo \cite{lipson2021raftstereo}, a learning-based method that estimates the stereo camera disparity and, thus, the scene depth. It is an extension of RAFT optical flow \cite{teed2020raft} and both methods are based on the core concept of recurrent iterative update and using all-pixel pairs to obtain cost volumes. RAFT-Stereo restricts the search problem to a linear unidirectional search as the search problem is limited along the epipolar line. The following is a brief description of Raft-Stereo \cite{lipson2021raftstereo} framework: Let $S_{L}$ and $S_{R}$ be the rectified stereo left and right images, respectively. Both $S_L$ and $S_R$ are passed through a feature encoder $F$, and one of the images $S_L$ is passed through a context encoder $C$, thus resulting in two feature vectors $F(S_L)$ and $F(S_R)$ and one context vector $C(S_L)$.  The feature vectors are used to make a correlation pyramid, which, along with the context feature vector, is provided to the \ac{gru} layers to iteratively estimate the disparity $D_{L,R}$ and thus depth image and point cloud $\mathcal{P}$. Let $\mathcal{P}_{t}$ be the pointcloud estimated at time $t$ and $\mathcal{P}_{t-1}$ be the pointcloud at time $t-1$. The disparity field starts with zero initialization, we augment the framework by replacing the naive zero-intialization with a warm start using a predicted disparity field using the last estimated point-cloud.

\subsection{Warm starting and filtering depth estimation}
\subsubsection{Warm start} We leverage the iterative framework of \cite{lipson2021raftstereo} and use the subsequently previous depth estimation to warm-start the iterations. To achieve this we use the past point-cloud $\mathcal{P}_{t-1}$ and use the $SE3$ pose transform (from \ac{reaqrovio}) between the current instance $t$ and the last depth image instance $t-1$ to transform the previous point-cloud $\mathcal{P}_{t-1}$ in the frame of the current time instance $t$. We then use \cite{katz2007direct} to remove the occluded points (also refered as hidden points) in the point-cloud. Lastly, we use the transformed past point-cloud to form first, the predicted depth image, and then, the predicted disparity image at time $t$.

\subsubsection{Filtering the depth image}
The estimated depth image is filtered to minimize the artifacts due to challenges in the vision. First, we obtain a confidence map by warping the feature vector of right camera image $F(S_R)$ using the disparity map $D_{L,R}$ and then we calculate the cosine similarity to obtain the confidence map. We reject all the points under a threshold of confidence and above a maximum distance.
\subsection{Mapping}
The filtered depth image is then used to project the points in 3D to obtain a point-cloud in the local robot frame. The point-cloud is then transformed in the inertial frame using the odometry (robot state estimates) from the \cref{sec:vision}. The transformed pointcloud in the world frame is then used to construct the map using the Voxblox mapping framework \cite{oleynikova2017voxblox}.

\section{Exploration and Inspection Path Planning}\label{sec:exploration}
In this work, we utilize our work described in~\cite{2023expgvi}, which is built on top of our open-sourced work on Graph-based exploration path planning (GBPlanner)~\cite{GBPLANNER_JFR_2020,GBPLANNER2COHORT_ICRA_2022}. The method was originally designed for the exploration and inspection of multiple compartments of ballast tanks, however, its extension to any given volume is straightforward.

The planner operates in two planning modes, namely \ac{ve} and \ac{gvi}. Operating on a volumetric representation of the environment, the first aims to explore and map a given enclosed volume $V$ with a depth sensor $\Ys_D$, whereas the latter finds a set of viewpoints and connecting path to inspect all surfaces of the mapped volume using a camera sensor $\Ys_C$ at a desired maximum viewing distance $r_{\max}$. We further assume that the robot utilizes a set $Y_S$ of camera sensors for \ac{vio}.
The planner starts in the \ac{ve} mode and iteratively plans paths to map the previously unknown volume $V$. In each iteration, the planner builds a $3$D graph $G_L$ within a local volume $B_L$ around the current robot location $\xi_0$ by sampling points in free space in $B_L$. In order to avoid the robot planning paths in fully open water (i.e., away from visible surfaces of the environment), the sampling method is modified to only retain samples having their distance to the closest occupied voxel less than a threshold $d_{obs}$. Next, the shortest paths in $G_L$, starting from $\xi_0$ to each vertex, are calculated. An information gain called the \textbf{VolumeGain} $\Gamma_D$, relating to the amount of volume seen, is calculated for each vertex in the graph, and the path having the highest accumulated $\Gamma_D$ is commanded to the robot. To guide the robot to maintain some visual features in the \ac{fov} of one of the cameras in $Y_S$ at all times, we only retain the the vertices for which $\eta$ (tunable parameter) percent of the combined \ac{fov} of the cameras in $Y_S$ sees occupied voxels in the map.

Upon completing the exploration, the planner switches to the \ac{gvi} mode. It builds a grid of viewpoints in $3$D at $r_{\max}$ from the mapped surfaces. The viewpoints are connected by collision-free edges to form a graph $G_{GVI}$ and additional points are sampled and added to $G_{GVI}$ to ensure no viewpoints are disconnected from $G_{GVI}$. The set of viewpoints providing complete coverage of the mapped surface is selected, and the path connecting them is calculated by solving the \ac{tsp}.

\begin{figure}
    \centering
    \includegraphics[width=1.0\linewidth]{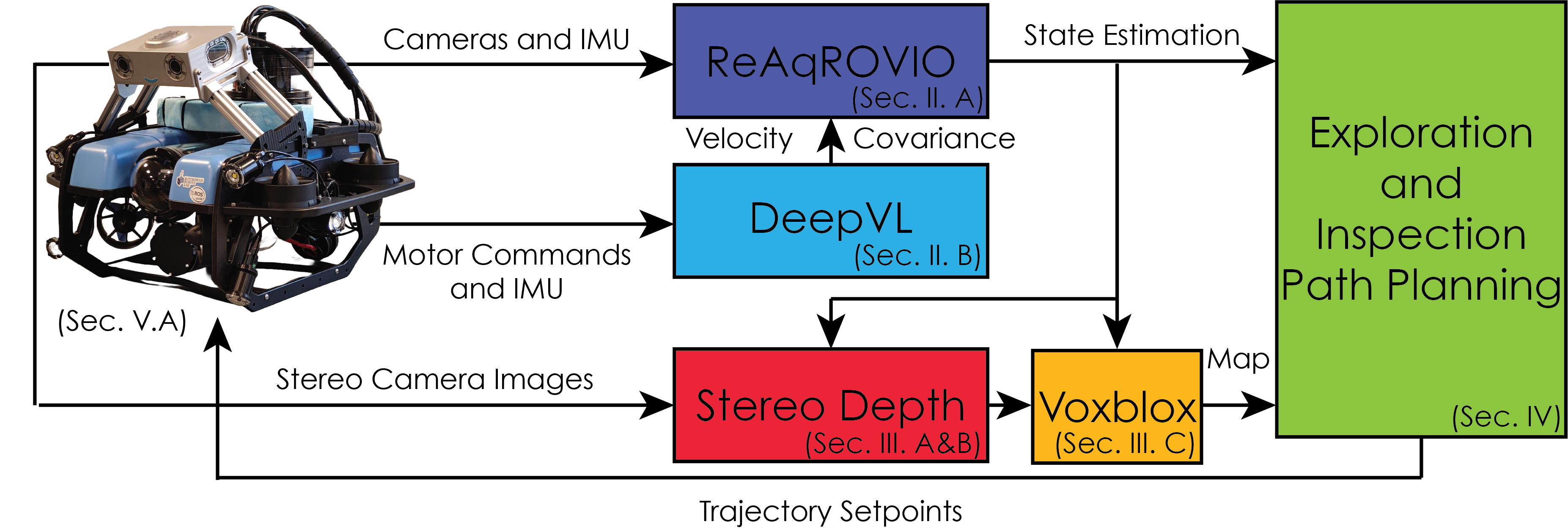}
    \caption{Overview of the software architecture of the autonomy stack running on Ariel.}
    \label{fig:enter-label}
\end{figure}

\section{Field Deployment}\label{sec:evaluation}
In this section, we present the details of the field experiment for the proposed vision-driven underwater exploration and inspection. A video of the experiments is available at: \href{https://ntnu-arl.github.io/ariel-explores/}{https://ntnu-arl.github.io/ariel-explores/}, and readers are encouraged to view it for more insights."

\subsection{Ariel: The custom underwater robot}
\begin{figure}
    \centering
    \includegraphics[width=1\linewidth]{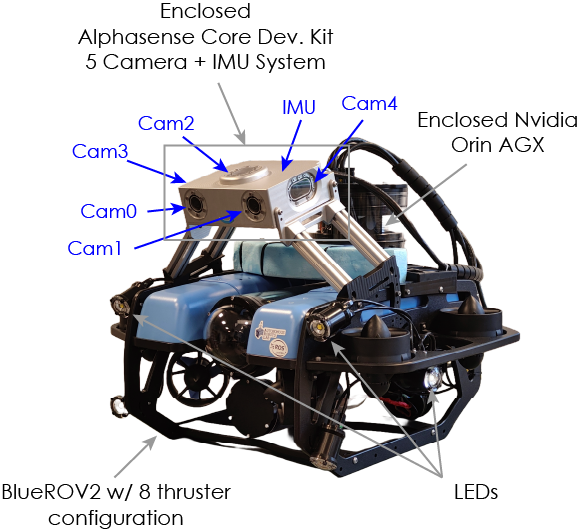}
    \caption{Sensing and compute setup of Ariel, the custom underwater robot. The robot contains an Alphasense core synchronized $5$ camera and \ac{imu} system as its main sensing suit, an Nvidia Orin AGX running the perception and planning software, and a Flir Blackfly S color camera. The robot is actuated by $8$ thrusters that enable 6 \ac{dof} actuation. A Pixhawk 6x flight controller is used as the low-level autopilot.}
    \label{fig:robot}
\end{figure}

We present a custom underwater robot, called Ariel, built on top of the BlueROV2 Heavy Configuration. It contains an Alphasense core system of $5$ synchronized cameras and \ac{imu}, a Blackfly S color camera, and an Nvidia Orin AGX 32 Gb Onboard computer running the online perception and planning software as shown in Figure~\ref{fig:enter-label}. The robot also houses a Pixhawk 6X flight controller and T-Motor C55A 8-in-1 electronic speed controller for the $8$ thrusters that enable 6 \ac{dof} actuation of the robot. The robot is fully powered by an onboard Lithium-ion battery. An illustration of the robot is shown in Figure~\ref{fig:robot}.

\begin{figure*}[ht]
    \centering    \includegraphics[width=0.96\linewidth]{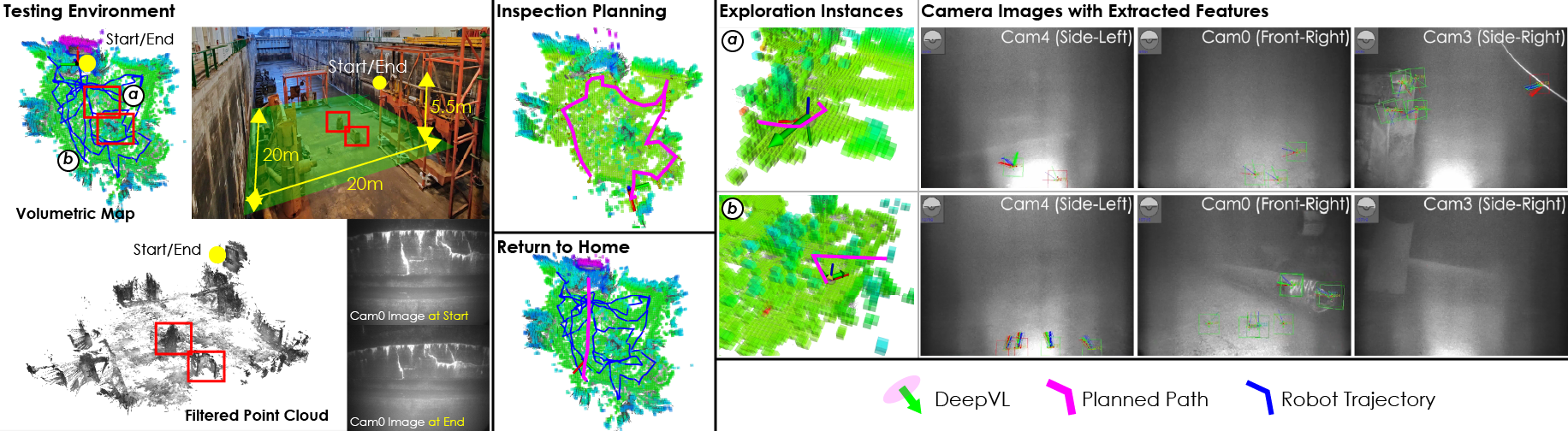}
    \caption{Results of field deployment in submarine bunker dry dock. The subfigures on the left show the final volumetric map with the robot trajectory and the filtered point cloud map. An image of the empty dry dock is presented to showcase the submerged structures and the dimensions. The onboard camera images at the beginning and the end of the mission shown are nearly identical which qualitatively demonstrates  the accuracy and robustness of the state estimation solution. On the right side, path planned during two instances of the \ac{ve} mode, the \ac{gvi} path, and the homing path are shown. Finally, the onboard camera images highlight the low visibility and the lack of visual features.}
    \label{fig:dora-result}
\end{figure*}
\subsection{Experiments in the submarine bunker}
The proposed framework was field-tested in a submarine bunker dry dock in the city of Trondheim, Norway. The environment had negligible external illumination due to the test being conducted in the late afternoon during dark Nordic winter. The visual conditions were further deteriorated by the use of unfiltered water from the Trondheim Fjord. 
These visual challenges result in a maximum visibility of $\approx\SI{2.5}{m}$. The dry dock was filled with \SI{6}{m} water and contained a dummy subsea industrial infrastructure as shown in Figure~\ref{fig:dora-result}. The robot was bound to remain in a volume of \(20\times20\times3\text{m}^3\).
The robot started close to a wall at a depth of \SI{1}{m} facing the wall. It was manually descended to the floor of the dock. The robot then initiated \ac{ve} mode while sampling points in the vicinity of the mapped regions. It took \SI{10}{min} to conclude the autonomous exploration with an average linear velocity of \SI{0.6}{m/s}. The exploration was then followed by \ac{gvi}. In this mode, the robot planned a single path to inspect the mapped surfaces with Cam0 (front right camera as shown in Figure~\ref{fig:robot}) at a maximum distance of $\SI{1.5}{\meter}$. Finally, the robot performed autonomous homing, returning to the starting point and ending the mission.

The visibility conditions can be assessed from the camera images shown in Figure~\ref{fig:dora-result}, showcasing the challenging visual conditions and lack of visual features. The sparseness of the filtered stereo depth pointcloud further highlights the low visibility in the water. 
Despite the challenging visual conditions, the proposed state estimation solution provided accurate odometry throughout the mission. This can be qualitatively assessed by comparing the camera images at the start and the end of the mission, which are nearly identical as the robot returns to the origin, i.e. the starting position before manual repositioning, as shown in Figure~\ref{fig:dora-result}.

\section{Lessons Learned}\label{sec:lessons}
\subsection{Robust Vision based Underwater State Estimation}

Underwater perception and state estimation are dominated in large part by acoustic sensing and vision is considered a secondary modality for perception. This is due to a multitude of challenges in using underwater vision, ranging from refractive effects (a varying camera model due to varying refractive index) to lack or blackout of visual features resulting in unreliable (or critical failure) state estimation. To this end, our solution tackles both these challenges by employing refraction-aware visual-inertial odometry and learning-based proprioception that bridges the state estimation in the scenario of deteriorated vision. This highlights the importance of methods such as DeepVL to aid the vision-based perception solution in situations of visual degradation. Furthermore, if robust, long-term, reliable state-estimation is achieved then autonomy from aerial robots can generalize to underwater robots.

\subsection{Generalist Path Planning Solutions}
The exploration and inspection path planning algorithms used in this work were originally developed and tested on aerial and legged robots. However, due to the generalist design of the method, the transfer to an underwater system was seamless. This highlights the benefit of modular autonomy architecture where certain components of autonomy, such as path planning, can be generalized and abstracted out from robot-specific components, like perception, to enable equivalent levels of autonomy across robot embodiments. 

\section{CONCLUSIONS}\label{sec:concl}
In this paper, we demonstrate aerial drone-level exploration and inspection autonomy on an underwater robot. A robot carrying a $5$ camera and imu sensing unit is presented along with the perception and path planning algorithms running onboard the robot. The system is deployed in an old inactive submarine bunker dry dock to inspect submerged infrastructure. The refraction-aware multi-camera visual-inertial state estimation method aided by a learning-based proprioceptive robot velocity prediction demonstrated accurate and robust state estimation despite of low visibility and low visual features. The path planning algorithms developed for aerial and ground robots are shown to seamlessly transfer to underwater robot enabling efficient exploration and inspection autonomy.
\newpage

\addtolength{\textheight}{-2cm}   

\bibliographystyle{IEEEtran}
\bibliography{BIB/main}

\end{document}